\pdfoutput=1

\documentclass[11pt]{article}

\usepackage[preprint]{acl}

\usepackage{times}
\usepackage{latexsym}
\usepackage{url}
\usepackage{CJKutf8}
\usepackage{booktabs}
\usepackage{amsmath}
\usepackage{amssymb}   
\usepackage{array}
\usepackage{subcaption}
\usepackage[ruled,vlined]{algorithm2e}
\usepackage{tikz}
\usepackage{tikzscale}
\usepackage{multirow}
\usetikzlibrary{bayesnet}
\usepackage{xcolor}
\usepackage{longtable}

\usepackage[T1]{fontenc}

\usepackage[utf8]{inputenc}

\usepackage{microtype}

\usepackage{inconsolata}

\usepackage{graphicx}

\title{Interpretable Mnemonic Generation for Kanji Learning \\ via Expectation-Maximization}

\author{
\bf Jaewook Lee, Alexander Scarlatos, Andrew Lan \\
  University of Massachusetts Amherst \\
  \texttt{\{jaewooklee,alexscarlatos,andrewlan\}@cs.umass.edu} \\
  }
  
\begin{document}
\begin{CJK}{UTF8}{min}
\maketitle
\begin{abstract} 
Learning Japanese vocabulary is a challenge for learners from Roman alphabet backgrounds due to script differences. Japanese combines syllabaries like hiragana with kanji, which are logographic characters of Chinese origin. Kanji are also complicated due to their complexity and volume. Keyword mnemonics are a common strategy to aid memorization, often using the compositional structure of kanji to form vivid associations. Despite recent efforts to use large language models (LLMs) to assist learners, existing methods for LLM-based keyword mnemonic generation function as a black box, offering limited interpretability. We propose a generative framework that explicitly models the mnemonic construction process as driven by a set of common \emph{rules}, and learn them using a novel Expectation-Maximization-type algorithm. Trained on learner-authored mnemonics from an online platform, our method learns latent structures and compositional rules, enabling interpretable and systematic mnemonics generation. Experiments show that our method performs well in the cold-start setting for new learners while providing insight into the mechanisms behind effective mnemonic creation.
\end{abstract}

\section{Introduction}

Learning vocabulary in a second language can be a cognitively demanding task, particularly when the writing system differs significantly from that of the learner’s native language. The challenge is especially pronounced in the context of Japanese, which employs not only hiragana, a syllabary roughly equivalent to the English alphabet, but also \textit{kanji}---logographic characters of Chinese origin. For learners whose native languages uses the Roman alphabet, mastering kanji is difficult due to their complexity and volume~\cite{everson2011best}. To ease this learning burden, keyword mnemonics~\cite{atkinson1975application} have been widely adopted as an effective learning approach. These mnemonics leverage the compositional structure of kanji; for example, the character 休 (rest) is composed of 人 (person) and 木 (tree), prompting the mnemonic ``a person resting by a tree,'' and learning platforms support this approach by assigning keywords, sometimes based on visual resemblance, to kanji components and offering cues that combine these keywords into memorable associations~\cite{heisig2011remembering,wanikani,kanshudo}.


\begin{figure}
    \centering
    \includegraphics[width=\linewidth, trim=2 2 2 2, clip]{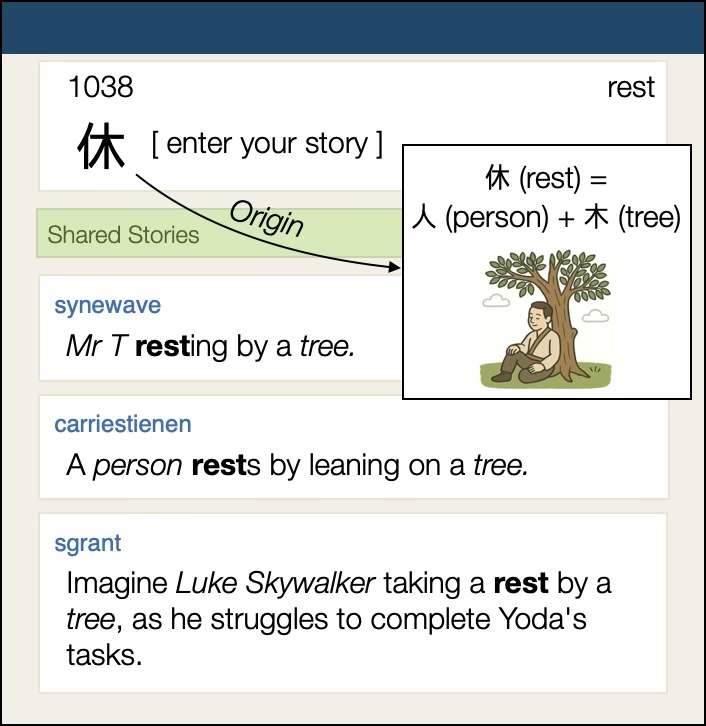}
    \caption{An example of an online platform \citet{koohii} for sharing mnemonics is kanji 休. A learner can type their mnemonic and share it with others.}
    \label{fig:rest_example}
\end{figure}

Recent work has explored automating the generation of such mnemonics using large language models (LLMs). \citet{lee2023smartphone} uses human-authored keywords in combination with LLMs to produce verbal cues and employs text-to-image models for visual aids. \citet{lee2024exploring} expands on this by allowing the LLM to generate both keywords and cues, followed by ranking mechanisms based on teacher feedback. \citet{balepur2024smart} applies supervised fine-tuning (SFT) and Direct Preference Optimization (DPO) to align generated mnemonics with learner preferences. \citet{kang2025phonitale} finds keywords in the learner’s first language by transliterating target-language phonemes into syllabic approximations, then uses these keywords with an LLM to generate verbal cues.

While these systems demonstrate the potential of LLMs to support language learning, they largely treat cue generation as a black-box process, offering less insight into how associations between keywords and target word are formed. This is in part because they rely on prompting or fine-tuning, which do not make the underlying mechanisms of cue construction transparent. Moreover, \citet{lee2024exploring}'s evaluation with language learners underscores the importance of accounting for individual variation in cue preferences. Learners rated cues based on imageability and coherence, and although LLM-generated cues scored comparably to human-authored ones on average, low inter-rater agreement indicated variability in individual preferences, highlighting the need for personalized cue generation approaches.

\textbf{Contributions}
To address the limited interpretability in existing LLM-based keyword mnemonic generation methods, we propose a generative framework that explicitly models the mnemonic construction process: there are a set of common \emph{rules} that learners follow when constructing mnemonics, and learners have different tendencies in using them for each kanji. We learn these rules and learner tendencies using a novel Expectation-Maximization (EM)-type algorithm. Rather than relying solely on LLMs, our approach learns latent structures and compositional rules from learner-authored mnemonics, enabling interpretable and systematic generation of mnemonics.




We evaluate our approach using data from an online language learning platform where learners share mnemonics for Japanese kanji. Since we focus on learning common rules among learners, rather than modeling the preferences of each individual learner, we mostly experiment in the cold-start setting, i.e., generating mnemonics for a new learner who has not created any mnemonics before. Results show that our EM-type method sometimes outperforms baselines, with modest but consistent gains in terms of generating mnemonics that better align with actual learner-authored ones. Furthermore, our analysis reveals interpretable rules and learner usage patterns, offering new insights into the mechanics of mnemonic generation.

\section{Problem Statement}
\label{sec:prob_stat}

We formally define and model the task of mnemonic generation, where learners author cues to help themselves memorize kanji. 
Assume we observe mnemonics authored by multiple learners for multiple kanji, denoted as
\begin{align*}
    \mathcal{D} = \left\{ \left(k_i, b_i, \{ m_{ij} \}_{j=1}^{J} \right) \right\}_{i=1}^I, (i,j) \in \Omega,
\end{align*}
where $j$ indexes learners, $i$ indexes kanji, \(k_i\) denotes the \(i\)th kanji, and \(b_i = \left( w(k_i), \{ w_i \} \right)\) denotes background metadata available to learners. For example, in online mnemonic platforms, there is typically some background on each kanji, such as information in language learning textbooks on the meaning and pronunciation of its components. Here, \(w(k_i)\) is the English meaning of the kanji, and \(\{ w_i \}\) is a set of associated keywords derived from its components. The set \(\{ m_{ij} \}_{j=1}^{J} \) contains verbal cues authored by different learners for kanji \(k_i\). Since in general, not all learners author mnemonics for all kanji, $\Omega \subseteq I \times J$. 

To promote learner participation in mnemonic authoring, which helps them learn kanji, our goal is to generate mnemonic cues for entirely new learners---i.e., a cold-start setting, in which no learner-specific mnemonics are available at test time. To model this, we ensure that all cues by a given learner are placed wholly in either the training, validation, or test set. Formally, letting $\mathcal{U}$ be the set of learners, we have:
\begin{align*}
    \mathcal U &= \mathcal U_{\mathrm{train}}\cup\mathcal U_{\mathrm{val}}\cup\mathcal U_{\mathrm{test}}.
\end{align*}
Our goal is to learn a model that captures common learner tendencies and preferences from mnemonics authored by learners in $U_{\mathrm{train}}$, to generate mnemonics for learners in the test set $U_{\mathrm{test}}$, i.e., 
\begin{align*}
    \hat{m}_{ij} \sim P(m_{ij} | k_i, b_i). 
\end{align*}

\section{Methodology}



We utilize an EM-type algorithm to jointly learn interpretable rules and latent variables that drive mnemonic generation. Our approach employs two language models: LM$_1$ as the mnemonic generator and LM$_2$ as the rule generator. We adopt a balanced configuration with LM$_1$ as the open-weights, trainable Llama-3.2 3B-Instruct model~\cite{meta2024llama32} and LM$_2$ as the proprietary GPT-4o~\cite{openai2024gpt4o}. It is crucial that LM$_1$ is open-weights because the E-step of our algorithm requires direct access to token-level likelihoods (details discussed below). We draw inspiration from \citet{shashidhar2024unsupervised}, who design modular systems where smaller LMs can be fine-tuned on learner-specific data without modifying the weights of an LLM. Similarly, our setup decouples mnemonic and rule generation, enabling lightweight adaptation without requiring end-to-end fine-tuning of the LLM. We note that our model choice was guided by capacity rather than licensing; the proprietary LLM is used solely to generate candidate rules from mnemonics, a role that could also be fulfilled by other large open-weight models.


\subsection{Latent Trait Model}

We model the process of writing a mnemonic \(m_{ij}\) as influenced by two latent variables: a learner-specific factor \(h_j\), which captures individual preferences, memory strategies, or stylistic tendencies; and a kanji-specific factor \(g_i\), which reflects semantic or structural features of the kanji that affect how keywords are interpreted or combined. 
We utilize a 1-Parameter Logistic Item Response Theory (1-PL IRT) model~\cite{rasch1993probabilistic} to estimate the probability that a mnemonic for kanji $k_i$ by learner $j$ involves rule $r_k$. In contrast to the traditional IRT formulation where item difficulty is subtracted from learner ability (i.e., $p = \sigma(\theta - \beta)$), we define the rule activation probability as:
\begin{align} \label{eq:irt}
p(z_{ijk} = 1) = \sigma(h_{jk} + g_{ik}),
\end{align}
where $h_{jk} \in \mathbb{R}$ denotes the affinity of learner $j$ for rule $r_k$, and $g_{ik} \in \mathbb{R}$ denotes the compatibility of kanji $k_i$ with rule $r_k$. Rule activations are given by $z_{ijk} \in \{0,1\}$, where $z_{ijk} = 1$ indicates that rule $r_k$ is used in generating mnemonic $m_{ij}$ for kanji $k_i$ by learner $j$. Therefore, in our formulation, high learner affinity ($h_{jk}$) indicates a personal tendency to use rule $r_k$; high kanji compatibility ($g_{ik}$) indicates that rule $r_k$ is semantically or visually suitable for kanji $k_i$.






\subsection{Learning Latent Traits via EM}
\label{sec:em}

\begin{algorithm}[t]
\caption{Our EM-type algorithm}
\While{not converged}{
  \tcc{E-step: Rule Assignment}
  \ForEach{pair \((i,j)\)}{
    \For{\(k = 1 \dots K\)}{
      \(p_k \gets P_{\text{LM}_1}(m_{ij} \mid b_i, r_k)\)
    }
    Let \(\mathcal{T}_{ij} \gets \text{Top-3 indices of } \{p_k\}\)\;
    \For{\(k = 1 \dots K\)}{
      \(z_{ijk} \gets \mathbb{I}[k \in \mathcal{T}_{ij}]\)
    }
  }
  \tcc{M-step: Update $h_{jk}$ and $g_{ik}$}
\vspace{-4ex}
  \[
  \mathcal{L} = \sum_{i,j,k} \text{BCE}\left(\sigma(h_{jk} + g_{ik}), z_{ijk}\right)
  \]
  Update \(\{h_{jk}\}, \{g_{ik}\}\) to minimize \(\mathcal{L}\)\;

  \tcc{Rule Update via \text{LM}\(_2\)}
  \For{\(k = 1 \dots K\)}{
    \(\mathcal{E}_k \gets \text{Top-N } \{ m_{ij}\} \text{ by } p_k\)\;
    \(r_k \gets \text{LM}_2(\mathcal{E}_k, \{r_{k'}\}_{k'\ne k})\)
  }
\ForEach{pair \((i,j)\)}{
  Fine-tune LM\(_1\) on \(P(m_{ij} \mid b_i, \{r_k : z_{ijk}=1\}\)
}
}
\label{alg:em}
\end{algorithm}

\subsubsection{Training}
After an initialization step we detail below in Section~\ref{sec:rule_init}, we utilize an EM-type algorithm to learn the latent learner traits $h_{jk}$ and kanji-specific compatibilities $g_{ik}$ that govern rule activation $z_{ijk}$. The whole process is summarized in Algorithm~\ref{alg:em}. 

In the E-step, for each kanji-learner pair $(i,j)$, we need to decide whether a rule $r_k$ is utilized when learner $j$ authors their mnemonic for kanji $i$. Therefore, we use the open-source $\text{LM}_1$ to compute the following likelihoods:
\begin{align} \label{eq:lik}
p_{ijk} \gets P_{\text{LM}_1}(m_{ij} \mid b_i, r_k), \quad \forall k.
\end{align}
In other words, we evaluate the likelihood that the learner-authored mnemonic $m_{ij}$ is generated by $\text{LM}_1$, while following rule $r_k$. However, we cannot completely relying on $\text{LM}_1$, which may not be fully calibrated to the mnemonics generation task, to select which rules are activated. Therefore, to make this process more robust, we set
\begin{align*}
z_{ijk} \gets \mathbb{I}[k \in \mathcal{T}_{ij}],
\end{align*}
where the $\mathcal{T}_{ij}$ are indices of the top-3 rules with highest per-token likelihoods of $m_{ij}$. Alternatively, one can set a likelihood threshold and select rules with likelihoods above this threshold; however, in our experiments, we find that such an approach is highly sensitive to the threshold parameter and less robust than taking the top-3, perhaps due to the high diversity in learner-authored mnemonics. 

%
In the M-step, we perform several different steps: First, we update the values of the latent variables $\{h_{jk}\}$ and $\{g_{ik}\}$, by minimizing the binary cross-entropy (BCE) loss between predicted activations $\sigma(h_{jk} + g_{ik})$ and observed values $z_{ijk}$, under the 1-PL IRT model, according to Eq.~\ref{eq:irt}. This step aligns the learner rule preferences and kanji rule compatibility with the rule activations $z_{ijk}$. 


Second, we update the rules $\{r_k\}$ by retrieving the most likely mnemonic set $\mathcal{E}_k$ under each rule, by selecting the top-8 mnemonics using previously computed values, $p_{ijk}$. We choose the number 8 because it falls within the range of the number of rules used in our later experiments. For rule updates, given $\mathcal{E}_k$ and the other rules involved in them (as decided in the E-step), we prompt LM$_2$ to generate a rule that is common in $\mathcal{E}_k$, but orthogonal to $\{r_{k'}\}_{k' \ne k}$ (Supplementary Material Table~\ref{tab:prompt_ortho}). This orthogonality encourages each rule to capture distinct semantic aspects of the mnemonics, reducing redundancy across the rule set. To promote orthogonality, we use updated rules for $k' < k$ and previous rules for $k' > k$ when generating $r_k$, ensuring that each new rule complements rather than overlaps with existing ones. 

We note that our method does not need a lot of API calls to the large, proprietary LM$_2$. Our method requires only $K \!\!\times \!\!T \!\!+ \!\!I \!\!\times \!\!J \!\!\times \!\!K$ calls to the proprietary LM ($T$ is the number of EM iterations): $K \!\!\times \!\!T$ during rule updates and $I \!\!\times \!\!J \!\!\times \!\!K$ for generating initial rule activations. It is a substantial reduction in proprietary LLM usage compared to fully fine-tuning LM$_2$, which requires $I \!\!\times \!\!J \!\!\times \!\!K \!\!\times \!\!T$ calls. 

Finally, we fine-tune LM$_1$ using the rule activations $z_{ijk}$ determined in the E-step with the newly updated $\{r_{k}\}$, along with the metadata for each kanji, $b_i$. Specifically, we maximize the log-likelihood of all mnemonics according to Eq.~\ref{eq:lik}, where we replace $r_k$ with the set of all rules relevant to the mnemonic, $\{r_k\}_{k \in T_{ij}}$. This setup further instruction-tune LM$_1$ to generate rules under instructions given by the updated rules, calibrating it for the mnemonic generation task. We then loop back to the E-step, recomputing likelihoods $p_k$ under the newly updated LM$_1$, repeating the process. This iterative procedure continues until early stopping is triggered based on validation loss. 


\subsubsection{Validation and Testing}
To support generalization to unseen learners and kanji in $\mathcal U_{\mathrm{val}}$ and $\mathcal U_{\mathrm{test}}$, we estimate trait values using population-level statistics derived from the training data. 
As we cannot estimate their individual traits $h_{jk}$ directly,  we use the mean learner trait vector $\bar{h}_k$ computed from training learners as a proxy. For each kanji $k_i$, if it appears in the training data, we use its learned trait $g_{ik}$ to compute rule activations using $\sigma(\bar{h}_k + g_{ik})$, and apply a threshold of 0.5 to determine whether each rule is active. Otherwise, we substitute the average trait $\bar{g}_k$ across all training kanji in place of $g_{ik}$. These activations guide LM$_1$ in generating or evaluating mnemonics during validation and testing. We apply early stopping based on validation loss, halting EM iterations once performance no longer improves.

\subsection{Interpretable Rule Initialization}
\label{sec:rule_init}
To kick-start the EM algorithm, we need to initialize the rules $\{r_k\}_{k=1}^K$ and rule activations $z_{ijk} \in \{0,1\}$. 
%
To bootstrap rule discovery, we sample 20 learners and collect their mnemonics to create a set of mnemonics, ensuring coverage among a diverse set of learners and kanji. We then utilize the proprietary LLM, LM$_2$, to generate an initial set of $K$ rules, by summarizing learner behavior when creating these mnemonics. 
We find that this step, despite using only a small set of mnemonics, can create a meaningful starting point of rules that correspond to broadly applicable mnemonic strategies, to be further optimized by our EM-type algorithm. 

Once the rules $\{r_k\}$ are initialized, we assign activations $z_{ijk}$ for each kanji-learner pair. We prompt LM$_2$ again to decide if each rule $r_k$ applies to the mnemonic $m_{ij}$. To promote sparsity in rule usage and avoid inclusion of marginally relevant rules, we prompt LM$_2$ to select at most three rules per mnemonic. We then initialize LM$_1$ by fine-tuning it to maximize the likelihood of mnemonic $m_{ij}$ given the input $b_i$ and the selected rules. 

\section{Experimental Settings}

\subsection{Dataset}

We evaluate our framework using data from \citet{koohii}, an online platform where Japanese language learners share mnemonics based on metadata from the book \emph{Remembering the Kanji}~\cite{heisig2011remembering}. We preprocess the data by selecting learners who provide mnemonic cues in English, yielding a dataset of 11,078 learners and 2,200 unique kanji, totaling 148,411 mnemonics (an average of 13.40 per learner). The book orders kanji by shared components (e.g., 集, 准, 進) and provides two types of metadata, $b_i$, per kanji: (1) a keyword together with a mnemonic supplied by the author and (2) a keyword only. To avoid bias from author-provided mnemonics, we retain only those kanji entries that include a keyword but no mnemonic. Finally, we filter the dataset using the OpenAI Moderation API~\cite{openai2025moderation} to remove toxic content.


\begin{table}[h]
\small
\centering
\begin{tabular}{lccc}
\toprule
\textbf{Set} & \textbf{\# Learners} & \textbf{\# Mnemonics} & \textbf{Avg. M/L} \\
\midrule
Train        & 323               & 8,991               & 27.84 \\
Val          & 40                & 1,164               & 29.10 \\
Test         & 40                & 1,156               & 28.90 \\
\bottomrule
\end{tabular}
\caption{Dataset split and summary statistics.}
\label{tab:dataset_stats}
\vspace{-.2cm}
\end{table}

We perform dataset splitting as follows: First, we filter out learners who authored less than five mnemonics. Then, we sort the remaining learners by the number of mnemonics they author, in descending order, and assign them to splits in an $8:1:1$ ratio (train:val:test) using a round-robin pattern. To control the dataset size, we uniformly subsample 25\% of learners from each split by selecting ones at fixed intervals across the sorted list, preserving the distributional diversity of the original dataset while reducing the volume for computational efficiency. Since the maximum length of a mnemonic is 180 tokens, we set maximum number of new tokens generated by LM$_1$ accordingly.

\subsection{Baselines and Metrics}


We define three types of baselines: zero-shot (\textbf{ZS}), supervised fine-tuning (\textbf{SFT}), and in-context learning (\textbf{ICL$_n^*$}). In the zero-shot setting, the base model is provided with metadata and prompted to generate a mnemonic without prior training. In contrast, the SFT baseline involves fine-tuning the model on a training set. The ICL$_n^*$ baselines serve as oracle comparisons, since they extend SFT by conditioning on $n$ example mnemonics from the same learner. These examples are not available to our method in the cold-start scenario, but they provide important cues about the learner’s preferences and stylistic tendencies. Thus, ICL$_n^*$ establishes an upper bound on what methods like EM can achieve without access to learner history. We report results for $n \in {1, 2, 3}$.


Our proposed method, \textbf{EM}, uses the same fine-tuning configuration as the SFT baseline, as described in Section~\ref{sec:imp}. The baselines use prompts similar to those in EM but without incorporating rules (See Supplementary Material Table~\ref{tab:prompt_for_gen}).

\subsubsection{Text Similarity Evaluation}
To evaluate how similar the generated mnemonics for a new learner is to the ground-truth, i.e., the mnemonic they author, we utilize \textbf{BERTScore}~\cite{zhang2019bertscore}, \textbf{ROUGE}~\cite{lin2004rouge}, and \textbf{LUAR}~\cite{uar-emnlp2021}. BERTScore measures semantic similarity between model outputs and human-authored mnemonics, while ROUGE measures lexical overlap. LUAR serves as an authorship verification metric, assessing whether generated cues resemble the writing style of learners in the test set. We also assess whether \textbf{Length} of the generated mnemonics matches the ground-truth, and report the ratio in length. 

\subsubsection{LLM-as-a-judge Evaluation}
\label{sec:llm_as_judge}
We utilize Prometheus evaluation~\cite{kim2024prometheus2} model for a head-to-head evaluation, comparing the mnemonics generated by our method against that generated by SFT, which mnemonic is more similar to the ground-truth, and measure the \textbf{Win Rate}. The comparison relies on a single in-context example from the learner's history to judge which mnemonic is more likely to have been authored by the learner (see Supplementary Material Table~\ref{tab:win_rate_prompt}).

We use ICL with one example for judging, for two key reasons: First, in cold-start cases with no learner history, even a single example can provide valuable insight into individual learner preferences. Second, because kanji are compositional, the previous kanji a learner wrote a mnemonic for is likely to share some keywords or components with the current kanji. Therefore, we use the learner's most recent mnemonic, and that kanji’s metadata, as the ICL example. 




\subsection{Implementation Details}
\label{sec:imp}
Each experiment was run using a single NVIDIA A40 GPU with 48 GB of VRAM. We utilize Low-Rank Adaptation (LoRA)~\cite{hu2021lora} to fine-tune LM$_1$, using a rank parameter $r = 128$, scaling factor $\alpha = 256$, and dropout rate 0.2. We set the learning rate to $1 \times 10^{-5}$. 
%
%
We empirically find that the number of rules to be $K = 10$ and later ablate the effect of varying $K$ in Section~\ref{sec:ablation}.

\section{Experimental Results}

We now detail both quantitative evaluation results and qualitatively analyze the learned rules and learner behavior patterns.

\begin{table*}[tp]
\small
\centering
\begin{tabular}{lccc|ccc|c|cc|c}
\toprule
\multirow{2}{*}{\textbf{Model}} & \multicolumn{3}{c|}{\textbf{BERTScore}} & \multicolumn{3}{c|}{\textbf{ROUGE}} & \multirow{2}{*}{\textbf{Length}} & \multicolumn{2}{c|}{\textbf{LUAR}} & \multirow{2}{*}{\textbf{Win Rate}}  \\
              & P & R & F1 & 1 & 2 & L & & CRUD & MUD & \\
\midrule
ZS$_\text{LM$_1$}$     & -0.119 & -0.048 & -0.084 & 0.135 & 0.017 & 0.093 & $\times$4.395 & - & - & - \\
ZS$_\text{LM$_2$}$     & -0.275 & -0.081 & -0.180 & 0.083 & 0.012 & 0.061 & $\times$10.08 & - & - & - \\ \midrule
SFT       & 0.212 & 0.005 & 0.105 & 0.249 & 0.047 & 0.181 & $\times$0.509 & 0.482 & 0.441 & Ref \\
EM & 0.240 & 0.009 & 0.121 & 0.251 & 0.048 & 0.183 & $\times$0.421 & 0.487 & 0.447 & 0.581 \\ 
ICL$_1^*$ & 0.192 & 0.066 & 0.128 & 0.259 & 0.051 & 0.187 & $\times$0.699 & 0.518 & 0.478 & 0.622\\
ICL$_2^*$ & 0.202 & 0.080 & 0.140 & 0.266 & 0.050 & 0.191 & $\times$0.696 & 0.522 & 0.486  & 0.637\\
ICL$_3^*$ & 0.207 & 0.089 & 0.147 & 0.274 & 0.054 & 0.197  & $\times$0.704 & 0.527 & 0.488 & 0.638\\
\bottomrule
\end{tabular}
\caption{Comparison across all methods on all metrics. ICL$^*$ uses prior mnemonics as in-context examples, thus establishes a performance upper bound on EM, which operates without having access to learner history.}
\label{tab:eval}
\end{table*}

\subsection{Overall Cold-start Performance}



Table~\ref{tab:eval} shows the evaluation results, comparing models on all metrics (when applicable). 
%
EM improves upon SFT in most aspects apart from length, with especially stronger gains on BERTScore and Win Rate.
According to Prometheus evaluation, using LLM-as-judge, EM is preferred over SFT with a win rate of 58.1\% vs.\ 41.9\%.
EM attains higher scores while providing greater interpretability in mnemonic generation compared to SFT.
The baseline models, ZS$_\text{LM$_1$}$ (Llama 3.2B) and ZS$_\text{LM$_2$}$ (GPT-4o), are prompted without any fine-tuning. These models perform poorly, generating outputs that are significantly longer than the ground-truth mnemonics. As a result, they receive low semantic and lexical scores and are excluded from further evaluation.
%
%


Our proposed method, EM, is designed for cold-start scenarios where no prior learner data is available. 
We see that ICL$_1^*$ slightly outperforms our method, and the gap between our method and this oracle is mostly smaller than the gap between our method and SFT, except for the LUAR metric. 
%
%
The reason is likely because while EM is designed for new learners, their style is still likely reflected in the mnemonics authored by other learners seen in the training set. Therefore, rules that we can learn can still anticipate some aspects of common learner behavior, and get close to the oracle performance. 
%
We prompt GPT-4o to evaluate whether the generated mnemonics adhere to the learned rules by identifying which activated rules are correctly applied. Compliance for each mnemonic is calculated by dividing the number of satisfied rules by the total number of applicable rules. Averaging these values across all examples yields an overall compliance rate of 60.58\%, indicating that while the EM-learned rules capture useful and generalizable patterns, their application during generation still leaves room for improvement.

%

Overall, our method offers a robust solution for generating high-quality mnemonics in practical settings, where no learner history is available. Next, we demonstrate the interpretability of our method by examining the rules it learns from learner-authored mnemonics. 


\begin{table*}[h]
\small
\begin{tabular}{cp{15cm}}
\toprule
\textbf{\#} & \textbf{Rules} \\
\midrule
1 & Emphasize transformation or conversion, illustrating how one element changes into another to clarify and connect the keywords with the kanji's meaning. \\
2 & Utilize simple, direct sentence structures focusing on clear actions or descriptions to establish a strong visual or conceptual link between the keywords and the kanji's meaning. \\
3 & Utilize familiar phrases or common knowledge to establish an intuitive and relatable link between the keywords and the kanji's meaning. \\
4 & Use logical reasoning or cause-and-effect relationships to connect the keywords with the kanji's meaning, providing a rational or explanatory narrative. \\
5 & Utilize cause-and-effect relationships within a single, direct sentence, emphasizing how the keywords lead to an outcome that encapsulates the kanji's meaning. \\
6 & Incorporate famous quotes or well-known expressions, either accurately or with a creative twist, to forge a memorable connection between the keywords and the kanji's meaning. \\
7 & Employ anthropomorphism, attributing human-like qualities or actions to non-human elements to forge a memorable connection between the keywords and the kanji's meaning. \\
8 & Utilize idiomatic or colloquial expressions, simplifying complex ideas into familiar sayings that evoke cultural references or visual imagery to link the keywords with the kanji's meaning. \\
9 & Utilize vivid imagery and sensory details to forge a memorable connection between the keywords and the kanji's meaning, engaging the reader's visual or sensory memory. \\
10 & Establish a direct association between the keywords and the kanji's meaning by framing them within a contextual scenario where their roles or functions naturally lead to the kanji's meaning. \\
\bottomrule
\end{tabular}
\caption{Top-10 learned mnemonic authoring rules from our EM-type algorithm.}
\label{tab:rules_k_10}
\vspace{-.2cm}
\end{table*}

\subsection{Learner and Kanji Clustering}

\begin{figure}[!htbp]
  \centering
  \includegraphics[width=.9\linewidth,trim={0mm 30mm 2mm 30mm},clip]{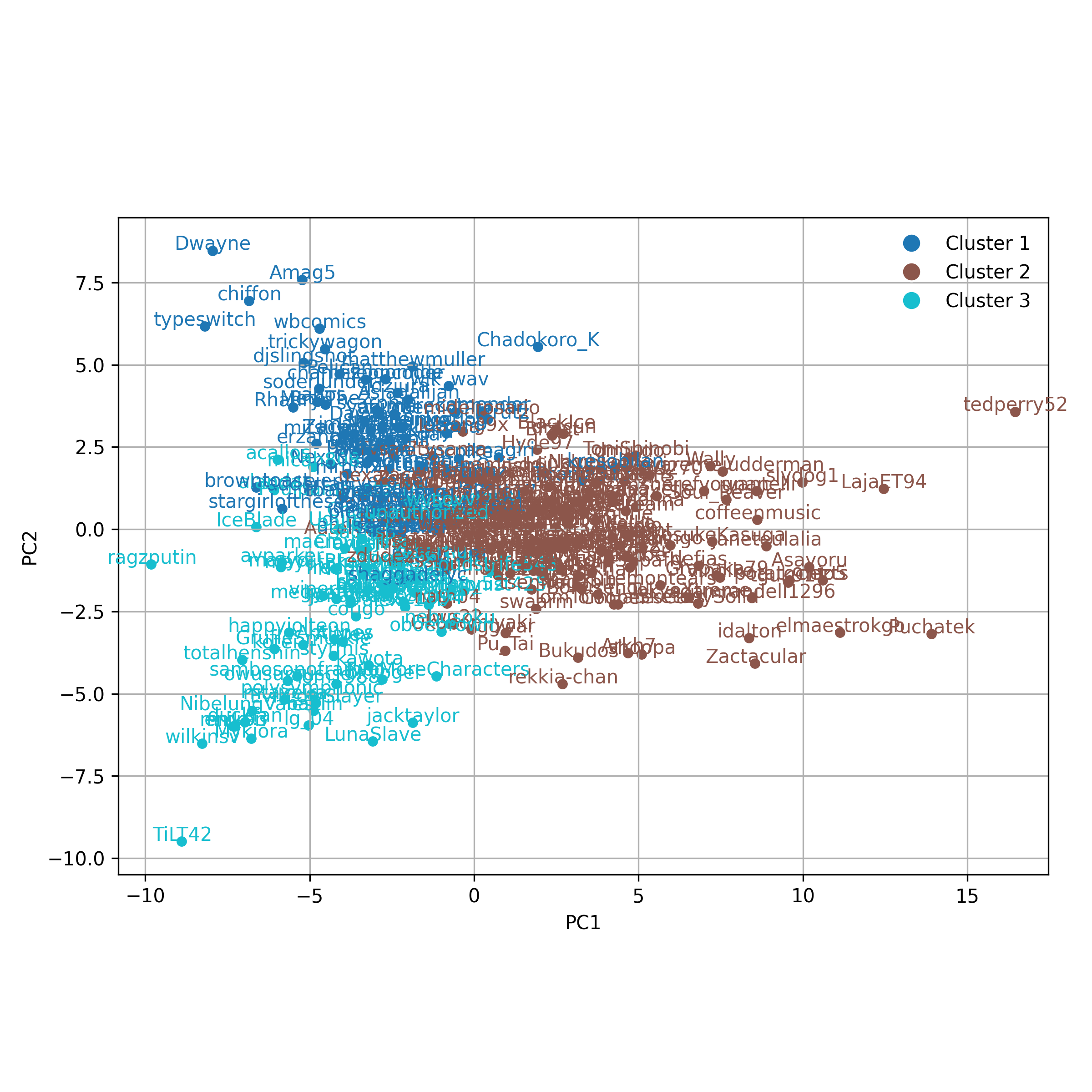}
  \caption{PCA plot for three learner clusters.}
  \label{fig:user_clusters}
\end{figure}
\begin{figure}[t]
    \centering
    \includegraphics[width=.9\linewidth,trim={0mm 30mm 0mm 30mm},clip]{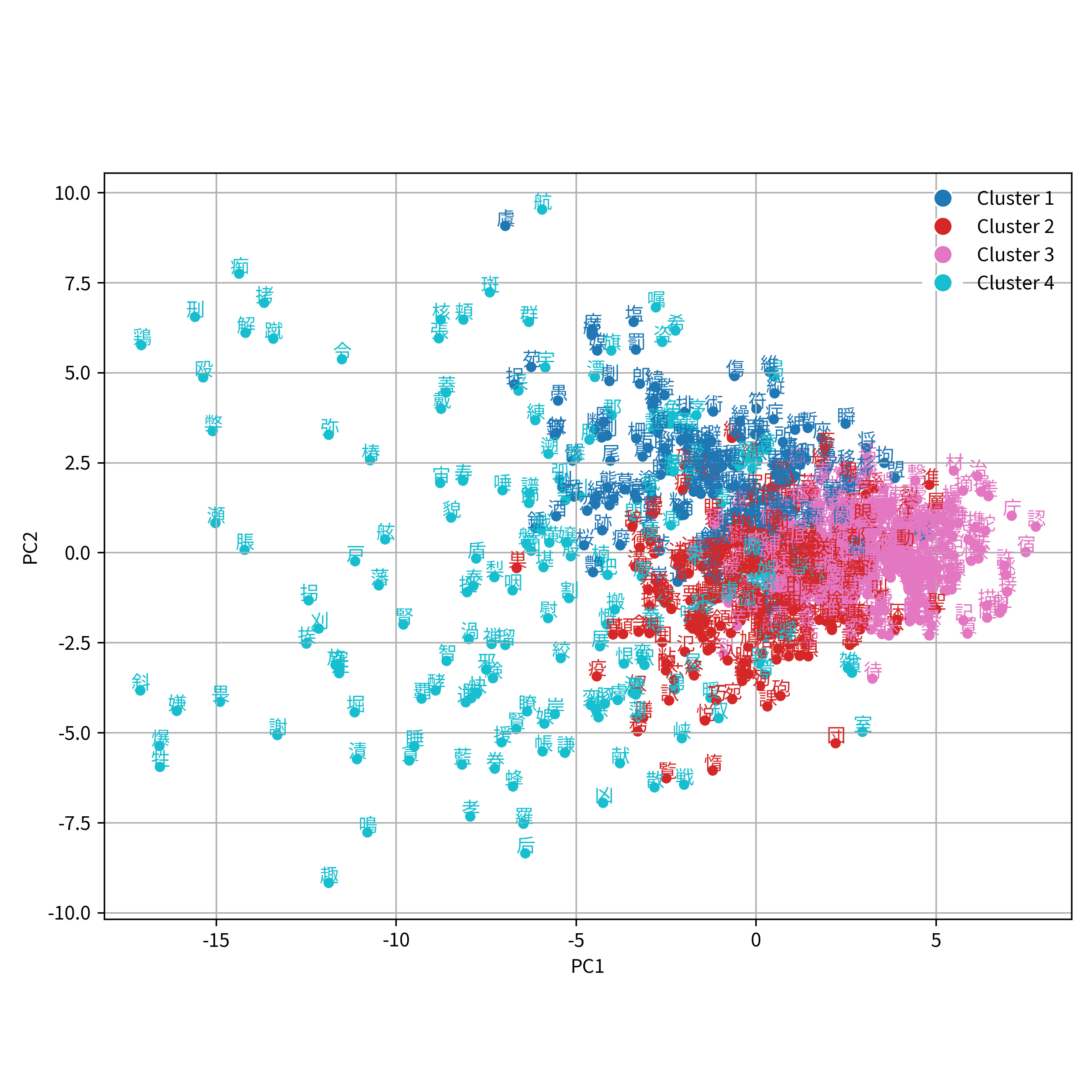}
    \caption{PCA plot for four kanji clusters.}
    \label{fig:kanji_clusters}
\end{figure}

To better understand how mnemonic rule preferences vary across learners and kanji, we perform unsupervised clustering on the learned latent learner rule affinity and kanji rule relevance parameters, $h_{jk}$ and $g_{ik}$. We use Gaussian mixture models and determine the number of clusters by selecting one with the maximum Bayesian information criterion~\cite{neath2012bayesian}. The clustering yields three learner clusters and four kanji clusters, shown in Figure\ref{fig:user_clusters} and Figure~\ref{fig:kanji_clusters}, respectively. The learner clusters contain 80, 1,711, and 72 learners, while the kanji clusters contain 193, 271, 369, and 178 characters.



To identify a representative learner or kanji for each cluster, we select the member whose rule usage vector is closest to the cluster centroid in the Euclidean space. Learners are grouped into clusters, each with its own centroid, and for each cluster, we select a representative learner as the one whose rule vector is nearest to that cluster's centroid. Similarly, kanji are grouped into clusters with corresponding centroids, and for each kanji cluster, the representative kanji is chosen as the one whose vector is closest to the cluster centroid.






We investigate cluster 2’s representative learner by examining the top three rules they most often use. First, the learner lean on common knowledge (Rule 3): for the kanji 團 ``group'' (keywords pent in + glued), their mnemonic is ``A group has to stick together,'' and for the kanji 届 ``deliver'' (keywords flag + sprout), their mnemonic mirrors the familiar quest-flag on game maps. Next, they use casual idioms (Rule 8): for the kanji 怠 ``neglect'' (keywords pedestal + heart), their mnemonic is ``If a person is put on a pedestal, they’ll eventually neglect the hearts of those who put them there,'' tying an abstract idea to a well-known saying. Finally, they employ anthropomorphism (Rule 7): for the kanji 批 ``criticism'' (keywords finger + compare), their mnemonic is ``The ring finger got a lot of criticism from the other fingers,'' giving non-human parts their own voice. These intuitive cues, vivid idioms, and playful personifications are highly representative of mnemonic-authoring behavior of learners in cluster 2. This observation suggests a way to elicit useful sources for mnemonic generation by using the rules in reverse, as prompts to uncover the kinds of associations a learner might naturally make. For example, Rule 3 (common knowledge) can prompt the question ``What everyday saying does this bring to mind?''; Rule 8 (casual idioms) might inspire ``What kind of casual phrasing feels natural here?''; and Rule 7 (anthropomorphism) could lead to ``Can you imagine this as a story with characters?'' Such questions help surface familiar reference points that guide more personalized mnemonic creation.


Moving on to kanji cluster 3's representative kanji, 任, along with its top three affinity rules. The kanji means ``responsibility'' 任 (keywords person + porter). There are 15 learner authored-mnemonics with the kanji. By examining the mnemonics, we can see how Rules 1, 2, and 5 are effectively applied. One learner uses transformation (Rule 1) by portraying Jesus as a porter who carries the sins of humanity, turning that burden into a sacred responsibility. Similarly, another learner presents Han Solo evolving from a smuggler to a responsible porter of cargo, reinforcing the meaning through character transition. A third learner introduces a moral consequence where the act of impregnation leads to the birth of a person and, therefore, responsibility, again showing transformation in action. For Rule 2, a learner provides a simple and direct sentence: ``A porter (壬) is a person (人) with the responsibility to take care of people's luggage,'' making the kanji’s components and meaning immediately clear. One learner also uses straightforward language in depicting a person boldly confronting a king with the demand for responsibility, while another concisely shifts responsibility from Aragorn to Frodo, emphasizing clarity. Rule 5 is prominent in another learner’s mnemonic, where Mr.\ T’s great power leads him to take on the responsibility of being a porter, showing a direct cause and effect. A learner reinforces this logic humorously by twisting the Spider-Man quote: ``With great porter, comes great responsibility,'' illustrating how the role itself leads to duty. Finally, a learner captures the same cause-effect relationship, noting that a pilot, by nature of being a ``person porter,'' inherently holds responsibility. Altogether, these mnemonics not only show creativity but also follow structured memory principles through transformation, clarity, and logical consequence.



\subsection{Qualitative Case Studies}




We analyze the top-5 and bottom-5 mnemonic predictions, ranked by their MUD scores, and try to gain deeper qualitative insights into how well our method performs on the mnemonic generation task (See Supplementary Material Table~\ref{tab:qual_anylsis}.) 

The top-5 cases show two main themes: when there is strong semantic alignment between the kanji and the associated keyword, rules are not activated; however, when the alignment is weak or abstract, the rules help bridge the semantic gap. For example, 将 (leader) associated with keywords of ``turtles'' and ``vultures,'' we can easily putting leader to either one of the animal. Similarly, 忌 (mourning) associated with ``snake'' and ``heart'', where we can easily get an idea of snake grief on a loss. When there is a weak connection,  仲 (go-between) consists of ``person'' and ``in'' was activated with familiar phrases or common knowledge, leading to using reference as ``man in black'',  ``Han and the Empire'', which is a mix of the movies (``The man in black was the go-between between Han and the Empire.''). The ground-truth also uses the Star wars references, ``Anakin, Jedi council'' where it shows the rule can bridge the gap between the meaning and the keywords (``Anakin was employed as a go-between between the Jedi council and Palpatine.''). Similarly, for 袋 (sack), one of the keywords is ``garment.'' One of three rules, based on familiar phrases or common knowledge, leads to a mnemonic like ``A sack is a garment for carrying things.'' The ground truth follows a similar concept, stating, ``A sack is a poor substitute for a garment.'' Therefore, the rules can help bridge semantic gaps in weakly aligned kanji–keyword pairings, enabling the prediction of plausible mnemonics.

In contrast, the bottom-5 cases show two main themes: a few cases correspond to learners have very distinctive styles. One example is a unique learner, who often includes additional vocabulary or grammatical details that are difficult to predict and now seen in any other learner. Therefore, since our method learns a certain number of rules that many learners use when they create mnemonics, distinct and uncommon styles may not be fully captured. For instance, in the mnemonic for 渡 (to cross), they add 渡す (わたす, ``to hand over''), which focuses on vocabulary usage rather than the kanji’s components, a style that is not found among other learners, complicating the prediction task.

Some low-scoring cases also result from different interpretations of keywords. For example, 化 (change), made of ``person'' and ``spoon,'' leads our method to generate ``The person who changed the world with his spoon was Jesus.'' However, one learner interprets ``spoon'' metaphorically, as in the cuddling position ``spooning,'' rather than the utensil. This interpretation departs from more literal or familiar interpretations, such as ``Jedi spoon'' or ``bent spoon,'' and highlights the difficulty of predicting mnemonics when learners use individual, unconventional ways to interpret keywords.

\subsection{Ablation Studies}
\label{sec:ablation}

\begin{table}[tp]
\small
\centering
\begin{tabular}{ccc|c}
\toprule
\multirow{2}{*}{\textbf{Variant}} & \multicolumn{2}{c|}{\textbf{LUAR}} & \multirow{2}{*}{\textbf{Win Rate}}  \\
              & CRUD & MUD & \\
\midrule
$K=5$ & 0.479 & 0.441 & 0.511 \\
$K=15$ & 0.479 & 0.440 & 0.569 \\ \midrule
Random & 0.479 & 0.439 & 0.548 \\
$\bar{h}_{jk}$ only & 0.485 & 0.443 & 0.605 \\
$g_{ik}$ only & 0.487 & 0.446 & 0.577 \\
\bottomrule
\end{tabular}
\caption{Ablation studies on the number of rules and methods for rule selection.}
\label{tab:abl}
\vspace{-.2cm}
\end{table}


We conduct ablation studies on two aspects: (1) the number of rules and (2) the method for rule selection. The corresponding results are presented in the bottom two rows of Table~\ref{tab:abl}. 

To assess the impact of the number of rules $K$, on mnemonic generation performance, we run the EM algorithm with 5 and 15 rules, in addition to 10. In both cases, the algorithm also converges after two iterations. The learned rules are shown in Supplementary Material Table~\ref{tab:rules_k_5} and Table~\ref{tab:rules_k_15}, respectively. With $K=5$, four of the selected rules overlap with those in the $K=10$ set, except for the rule concerning the juxtaposition of unrelated or contrasting keywords. In contrast, the $K=15$ setting captures all 10 original rules and further subdivides them into finer-grained rules. Balancing performance and computational cost, we find that using $K=10$ offers a favorable trade-off.

To evaluate whether the learned latent variable generalizes to unseen learners, we modify rule activation at test time while keeping EM-learned parameters fixed. First, we randomly activate three rules during inference. Second, we activate rules based solely on $\bar{h}_{jk}$ resulting in only Rule 3 being used. Lastly, we activate rules using only $g_{ik}$. While randomly activating results in lower performance than activating $\sigma(\bar{h}_{jk} + g_{ik})$, activating rules by $\bar{h}_{jk}$ or $g_{ik}$ achieves competitive performance. These findings suggest both $\bar{h}_{jk}$ and $g_{ik}$ contribute meaningfully to rule activation. For future work, an important avenue is to investigate how to estimate $h_{jk}$ for new learners, possibly by efficiently updating it from limited examples or preference surveys before they start authoring mnemonics.

A central feature of our approach is the use of LM$_1$ within a rule-constrained generation pipeline. Rather than depending on large, black-box models for unconstrained generation, we constrain LM$_1$ to operate under rule guidance. This design enhances interpretability, enforces structural consistency, and permits fine-grained control over output. The small footprint of LM$_1$ also supports iterative updates across EM, enabling continual refinement of rule representations without high computational cost.

\section{Conclusion and Future Works}
In this paper, we explored the use of an Expectation Maximization-type algorithm to learn latent structures and compositional rules from learner-authored mnemonics. In a cold-start scenario with no learner history, our method sometimes outperformed baselines, generating mnemonics that better aligned with those authored by actual learners. We also showed that the learned interpretable rules and usage patterns revealed new insights into the mechanics of mnemonic generation.

There are many avenues for future work. First, in non cold-start settings, we can investigate how to build the rules we learned into a learner's persona \cite{shashidhar2024unsupervised} for better personalization. Second, we can explore how to leverage not just information on kanji components, but its structural information~\cite{yu2024chinese}, and incorporate it into mnemonic generation. Third, we can explore creating an interactive learning environment and use our work to aid learners during mnemonic creation, in a learner-AI teaming setting. 








\section*{Acknowledgments}
The authors are partially supported by the NSF under grants 2237676 and 2341948.

\section*{Limitations}
The study has a few limitations that should be considered. First, we evaluate mnemonic quality using a large language model rather than human raters. While automated evaluations allow for scalable and consistent comparisons, they may not fully capture the subjective qualities present in real-world learning scenarios. Incorporating human evaluations would provide a more comprehensive assessment.
Second, we explore our method using only the Llama 3.2-Instruct (3B) model. While this model offers a strong balance between performance and efficiency, results may differ with larger models or with other models of similar parameter size. Exploring a broader range of models could strengthen generalizability.
Third, our approach is tested primarily on a dataset of learner-authored mnemonics for kanji. While kanji is a natural case study due to its compositional structure, further work is needed to assess whether the proposed rule-based generative framework generalizes to other scripts or learning domains.
Finally, although our model accounts for variation in mnemonic construction styles across learners, it assumes that each learner's preferences are relatively stable and can be inferred from their past mnemonics. In practice, learner behavior may evolve over time or vary by context, suggesting that future work could explore adaptive or dynamic personalization strategies.

\section*{Ethics Statement}
We make use of the website Kanji Koohii, a learner-driven platform that provides mnemonic aids and community-contributed stories to support the learning and memorization of Japanese kanji. All learner-contributed content on the site’s Study pages is licensed under a Creative Commons Attribution-NonCommercial-ShareAlike (CC BY-NC-SA) license, which permits reuse and adaptation with attribution, for non-commercial purposes, and under the same license terms. Accordingly, we have credited the authors of the mnemonics by citing their usernames in this paper. The dataset itself is not redistributed here; only derived analyses and results are presented.

\bibliography{acl_latex}

\appendix

\onecolumn
\clearpage
\section*{\Large Supplementary Material}
\label{sec:supp}

\section{Related Work}
\label{sec:related_work}


\subsection{Keyword Mnemonics}
The application of keyword mnemonic generation using LLMs began with \citet{lee2023smartphone}, which used human-authored keywords to generate both verbal and visual cues, followed by a human evaluation to assess how variations in the cues impact learning. This work was extended by \citet{lee2024exploring}, who employed LLM prompting for both keyword and verbal cue generation, using a ranking system based on feedback from English teachers on mnemonic quality. \citet{balepur2024smart} further advanced the field by applying supervised fine-tuning and direct preference optimization to align generated mnemonics with learner preferences. It is worth noting that \citet{lee2023smartphone} focused on English-to-German mnemonics, while \citet{lee2024exploring} and \citet{balepur2024smart} targeted English-to-English mnemonics. 

\subsection{LLM Personalization}

In terms of LLM personalization, there are two main approaches: model-centric and data-centric.

\subsubsection{Model-Centric}

This approach adapts large language models to individual users via parameter updates, including both full fine-tuning and parameter-efficient tuning using LoRA and adapters. \citet{zhong2021useradapter} uses prefix-tuning to inject user information, introducing minimal overhead. \citet{zhang2024personalized} adapts LLMs by injecting low-rank matrices into the model's weights, employing a plug-and-play adapter framework that allocates a small LoRA module for each user. \citet{tan-etal-2024-personalized} introduces a collaborative mechanism by sharing parameter shards across users, enabling the construction of new personalized modules without retraining. \citet{qi2024fdlora} applies federated learning to maintain privacy, separating global and personal LoRA modules across devices. \citet{doddapaneni-etal-2024-user} uses fixed embeddings as a personalization strategy.

\subsubsection{Data-Centric}

\citet{zhang-etal-2018-personalizing} provides each dialogue agent with a profile (a ``persona'') in natural language. Conditioning on these profiles makes responses more consistent and engaging, as the chatbot’s outputs stay aligned with the given persona traits. Modern LLMs extend this idea by allowing a system prompt or context that lists the user’s preferences, history, or persona traits. Rather than using static profiles, another input-based method includes personalized exemplars or dialogue history in the prompt. To efficiently personalize without overloading the prompt, many systems use a retriever to fetch the most relevant pieces of user data on the fly. In the LaMP benchmark, \citet{salemi2024lamp} augment user queries with the top-$k$ retrieved items from that user’s history, such as past posts or interactions, before feeding them into the LLM. \citet{ning2024user} uses latent embeddings for personalization, constructing dense user vectors by pretraining on a user’s interaction data, such as movie ratings and review texts, so that the vector encodes the user’s tastes. \citet{zhang2024personalized} also explores training persona embeddings or persona memory networks, which are infused into generation models to modulate their outputs.

\newpage
\section{EM algorithm}
\subsection{Rule Initialization}
Table~\ref{tab:init_rules} shows the initial rules generated from randomly sampled 20 learners from \citet{koohii}.
Algorithm~\ref{alg:init} presents the procedure for initializing the rules of $z_{ijk}$ using LM$_2$, as described in Section~\ref{sec:rule_init}.

\begin{table*}[h]
\small
\begin{tabular}{cp{15cm}}
\toprule
\textbf{\#} & \textbf{Rules} \\
\midrule
1 & Create vivid stories involving the keywords to form a memorable image associated with the kanji meaning. \\
2 & Use wordplay or puns related to the kanji's pronunciation to enhance recall. \\
3 & Incorporate familiar cultural references or media to make the mnemonic more relatable and memorable. \\
4 & Leverage the literal and metaphorical meanings of the keywords to construct a logical sequence or scenario. \\
5 & Form associations between the kanji and its meaning by imagining interactions or actions involving the keywords. \\
6 & Use exaggeration or humorous situations to make the mnemonic more engaging and memorable. \\
7 & Identify and emphasize the emotional states or feelings that relate to the kanji's meaning within the mnemonic. \\
8 & Utilize contrast or unexpected outcomes in the story to create a more striking mental image. \\
9 & Create mnemonics that involve transformation or change related to the keywords to connect with the kanji's meaning. \\
10 & Incorporate simple, straightforward associations for ease of recall, emphasizing clarity over complexity. \\
\bottomrule
\end{tabular}
\caption{Initial rules are generated using the stories from the following username and kanji: theadamie (批), vvrk79 (蝶), potatochip5 (泡), Joyo1945 (怖), BlackIce (場),  danschub (刷), nath04 (允), matthewmuller (憶), picassoisahack (錮), polysymphonic (誰), strugglebunny (蛇), Tvaw73 (詳), SketchySolid (起), Artemisk (逐), Bokusenou (萤), Chadokoro\_K (記), proagg (規), ragzputin (學), Eklmejlazy (摩), zdude255(進)
}
\label{tab:init_rules}
\end{table*}

\begin{algorithm}[H]
\caption{Rule Initialization}
\begin{minipage}{\linewidth}
  \tcc{Rule Discovery}
  Sample $M$ users $\{j_1, \dots, j_M\}$\;
  \ForEach{kanji $i$}{
    $\mathcal{C}_i \gets \{m_{i',j_m} \mid i' \ne i, j_m \in \{j_1, \dots, j_M\}\}$\;
  }
  Use LM\(_2\) to summarize $\{\mathcal{C}_i\}$ into $\{r_k\}_{k=1}^K$\;

  \vspace{1ex}
  \tcc{Rule Activation}
  \ForEach{pair $(i,j)$}{
    \For{$k = 1 \dots K$}{
      $z_{ijk} \gets \mathbb{I}[\text{LM}_2 \text{ aligns } r_k \text{ with } m_{ij} \text{ given } b_i]$\;
    }
  }

  \vspace{1ex}
  \tcc{Language Model Conditioning}
  \ForEach{pair $(i,j)$}{
    Fine-tune LM\(_1\) on $P(m_{ij} \mid b_i, \{r_k : z_{ijk} = 1\})$\;
  }
\end{minipage}
\label{alg:init}
\end{algorithm}

\newpage
\subsection{Learning Latent Trait via EM}
Table~\ref{tab:prompt_ortho} shows the prompt for generating orthogonal rules, as described in Section~\ref{sec:em}.

\begin{minipage}{\linewidth}
  \centering
  \begin{tabular}{p{15cm}}
    \toprule
    \textbf{Task} \\
    Your task is to generate a new rule based on mnemonic stories written by users. \\
    Focus on identifying \textbf{common patterns} or \textbf{recurring techniques} across multiple stories, \\
    but ensure that the rule you create is \textbf{orthogonal} (i.e., distinct and non-overlapping) \\
    to the existing rules provided below. \\
    \textbf{Existing Rules} \\
    Below are other rules that have already been established: \\
    - Rule 1: \textit{(rule here)} \\
    - Rule 2: \textit{(rule here)} \\
    $\cdots$ \\
    \textbf{New Input Examples} \\
    --- \\
    Kanji (Meaning): \textit{example kanji} (\textit{example meaning}) \\
    Keywords: \textit{keyword1, keyword2, keyword3} \\
    Story: \textit{example mnemonic story here} \\
    --- \\
    \textbf{Output Format} \\
    Use \texttt{<thinking></thinking>} to explain the common patterns or narrative elements you observed \\
    across the stories. Then, propose a new rule enclosed in \texttt{<rule></rule>} that captures these shared elements. \\
    The new rule must be \textbf{one clear, single sentence} and must not overlap with the existing rules listed above. \\
    \textbf{Output} \\
    \\
    \bottomrule
  \end{tabular}
  \captionof{table}{Orthogonal rule generation prompt used in Algorithm~\ref{alg:em}.}
  \label{tab:prompt_ortho}
\end{minipage}


\newpage
\section{Learned Latent Variables}
\label{sec:latent_variable}
\subsection{Heatmap}
\begin{figure*}[htbp]
    \centering
    \begin{minipage}{\textwidth}
        \centering
        \begin{subfigure}[b]{0.32\textwidth}
            \centering
            \includegraphics[width=\textwidth]{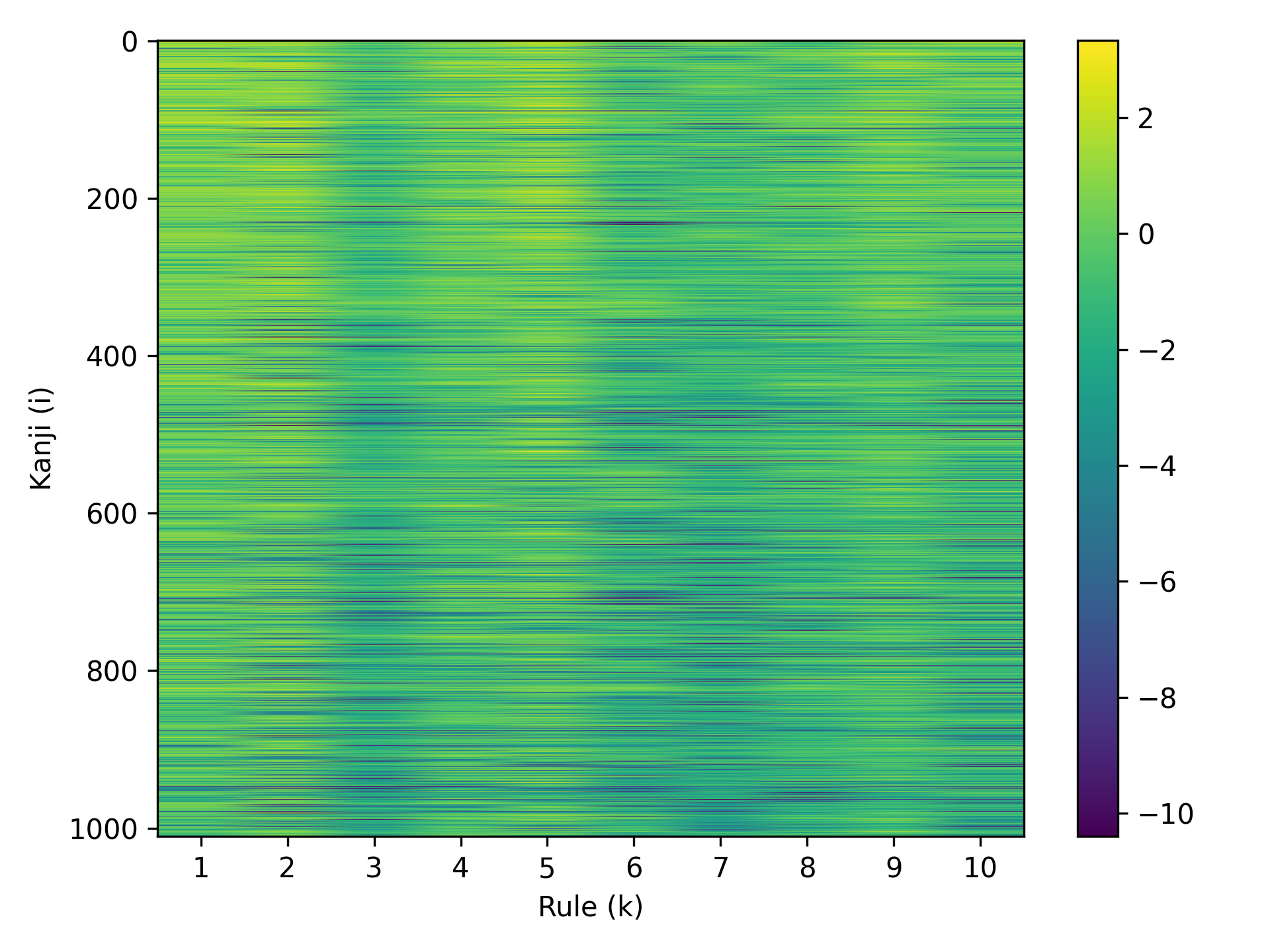}
            \caption{$g_{ik}$ (train)}
            \label{fig:heatmap_g_ik_train}
        \end{subfigure}
        \begin{subfigure}[b]{0.32\textwidth}
            \centering
            \includegraphics[width=\textwidth]{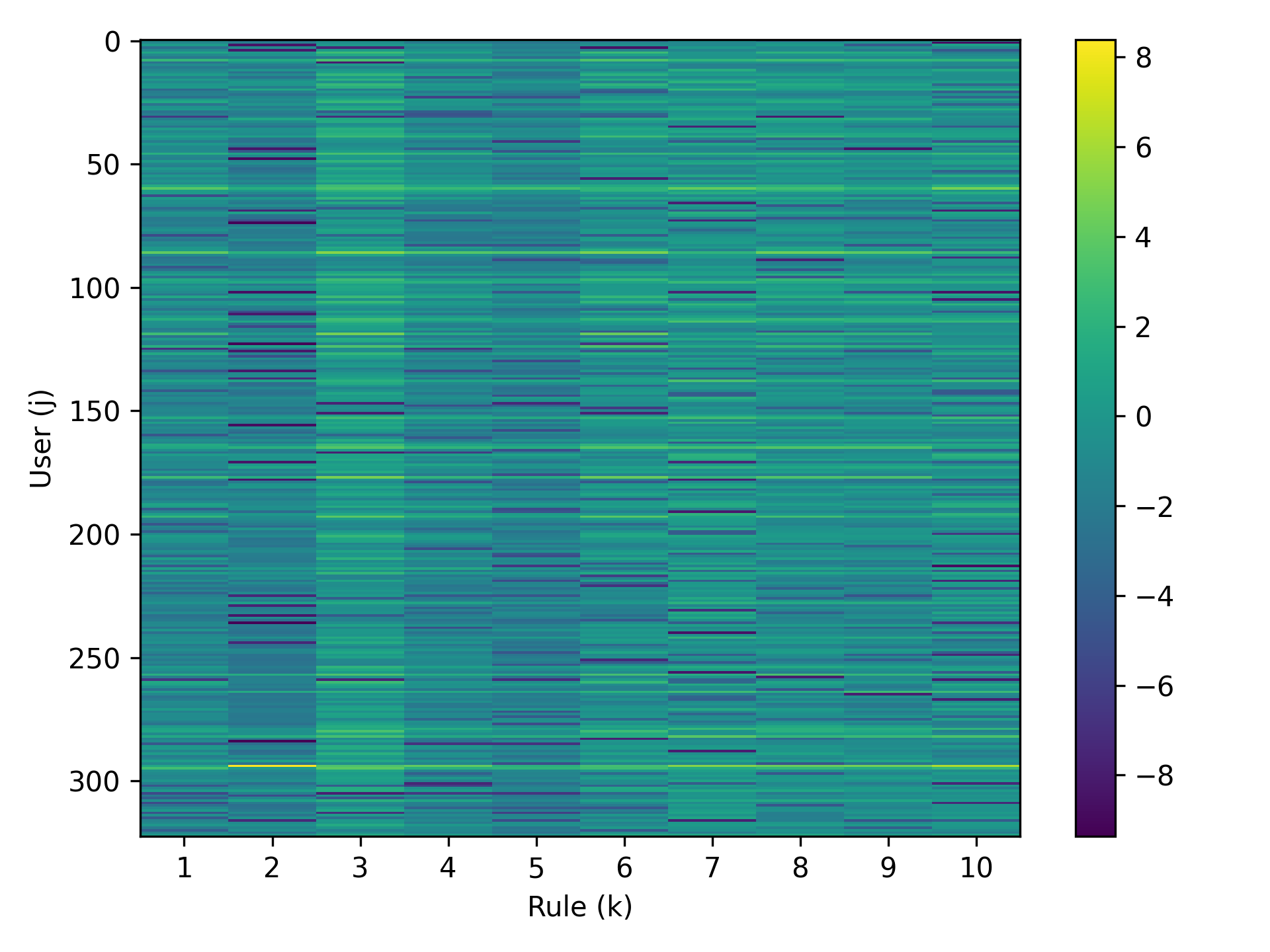}
            \caption{$h_{jk}$ (train)}
            \label{fig:heatmap_h_jk_train}
        \end{subfigure}
        \begin{subfigure}[b]{0.32\textwidth}
            \centering
            \includegraphics[width=\textwidth]{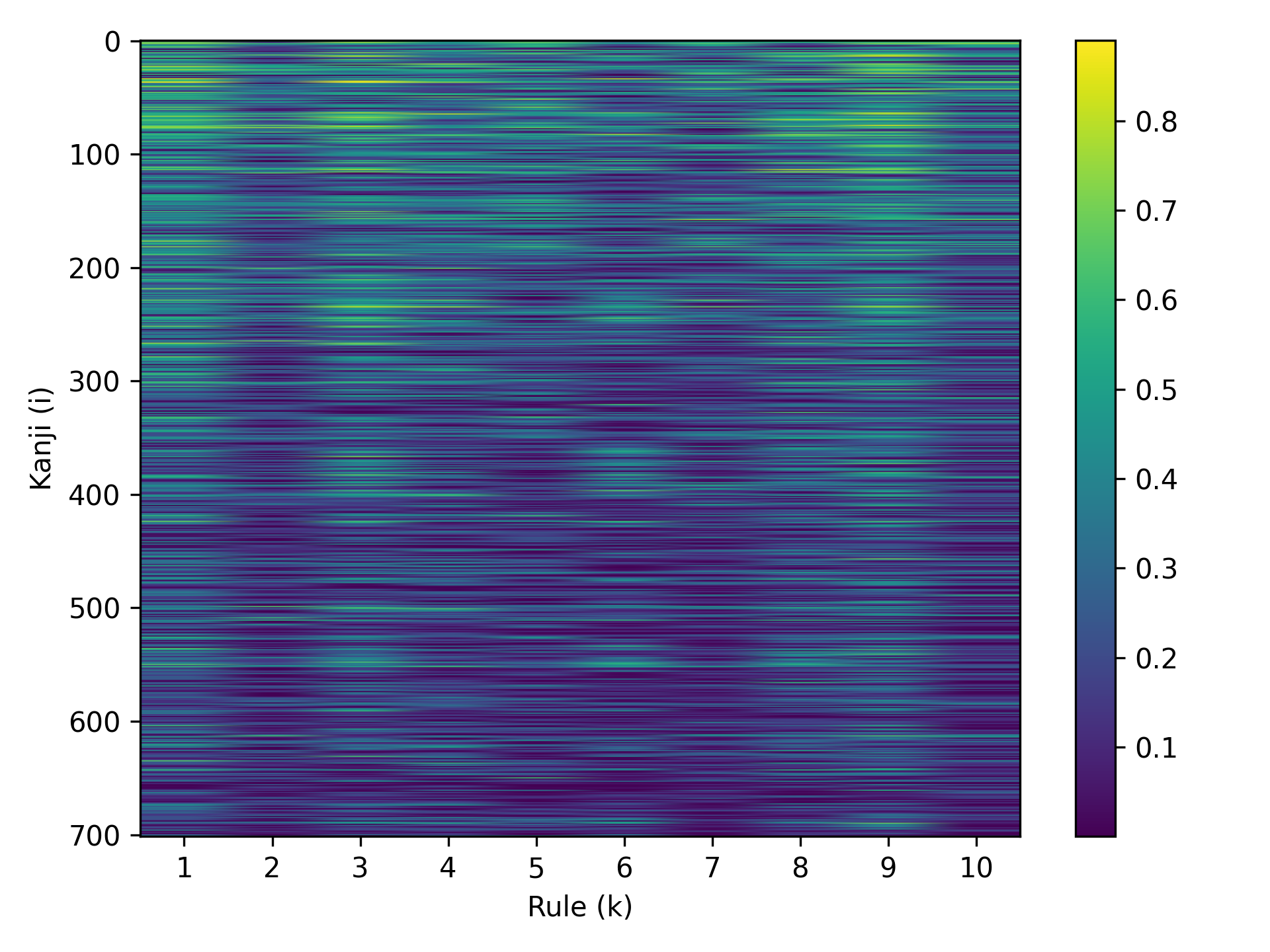}
            \caption{$\sigma(\bar{h_{jk}} + g_{ik})$ (test)}
            \label{fig:heatmap_sigmoid_hg_ik_test}
        \end{subfigure}
        \caption{Training and test heatmaps: the first two plots show the individual components $g_{ik}$ and $h_{jk}$ computed from the training set. The third plot shows the activated rules on the test set using $\sigma(\bar{h_{jk}} + g_{ik})$, where $\bar{h_{jk}}$ is the mean of $h_{jk}$ over training, and if the kanji $i$ doesn't exist in the training we use the average $\bar{g_{ik}}$.}
        \label{fig:all_heatmaps}
    \end{minipage}
\end{figure*}

\subsection{Cluster}
Three learner clusters sizes of 80, 1711 and 72 and four kanji clusters, 193, 271, 369, and 178.


\begin{table}[h]
\centering
\begin{minipage}{\textwidth}
\centering
\small
\begin{tabular}{cccccccccccc}
\toprule
\multirow{2}{*}{\textbf{Latent}} & \multirow{2}{*}{\textbf{Cluster}} & \multicolumn{10}{c}{\textbf{Rules}} \\
\cmidrule(lr){3-12}
                        &                          & 1 & 2 & 3 & 4 & 5 & 6 & 7 & 8 & 9 & 10 \\
\midrule
\multirow{3}{*}{$h_{jk}$}                   & 1                        & -2.343 & -1.733 & -1.691 & -2.340 & -2.513 & -2.533 & \textbf{-0.340} & -1.897 & \textbf{-1.393} & \textbf{-0.494} \\
                        & 2                        & -0.251 & -0.896 & \textbf{1.140}  & -0.232 & -0.728 & 0.393  & \textbf{0.519}  & \textbf{0.404}  & 0.142  & 0.116  \\
                        & 3                        & -1.826 & -4.195 & \textbf{-0.135} & -1.780 & -2.794 & \textbf{-0.757} & -2.918 & \textbf{-0.928} & -1.684 & -3.404 \\ \midrule
\multirow{4}{*}{$g_{ik}$}                    & 1                        & \textbf{-0.128} & -3.056 & -1.237 & \textbf{-0.544} & \textbf{-0.519} & -1.214 & -2.037 & -1.256 & -0.970 & -2.665 \\
                        & 2                        & \textbf{-0.344} & \textbf{0.371}  & -1.639 & -1.201 & -1.198 & -2.113 & -1.373 & -1.204 & \textbf{-0.875} & -1.032 \\
                        & 3                        & \textbf{0.583}  & \textbf{0.613}  & -0.862 & 0.155  & \textbf{0.652}  & -0.621 & -0.937 & -0.579 & 0.211  & -0.354 \\
                       & 4                        & \textbf{-1.921} & -2.618 & -4.361 & \textbf{-2.312} & -2.174 & -3.861 & -4.119 & -3.141 & \textbf{-1.778} & -2.877 \\
\bottomrule
\end{tabular}
\caption{Mean rule affinities per cluster and top-3 highest affinities are bolded.}
\end{minipage}
\end{table}

\subsection{Representative of Clusters}

\begin{minipage}{\textwidth}
\small
\centering
\begin{tabular}{l p{15cm}}
\toprule
\textbf{Kanji} & \textbf{Mnemonic} \\
\midrule
蛇  & Only Snake from MGS3 could survive inside of his cardboard house with nothing but a spoonful of insects. \\
鮮  & The F of Fresh comes from Fish and the SH from Sheep. \\
団  & A group has to stick together. \\
庁  & I imagine that King Bowser’s government office is like a cave with a lot of spikes. \\
批  & Ever since the ring was bestowed on the ring finger, it got a lot of criticism from the other fingers since, compared to the index finger, the ring finger isn’t very useful. \\
研  & Geodude is a rock Pokémon with two hands that can learn the move ``Rock Polish''. \\
淫  & Remember those lewd 80s porn tapes that all start with some porter entering a woman’s room who latches onto him like a vulture? It all just turns into a wet mess where they’re covered in each other’s fluids. \\
怠  & If a person is put on a pedestal, they’ll eventually neglect the hearts of those who put them there. \\
冶  & They say an advanced method for metallurgy is to use an ice-powered pedestal for the yet un-cooled metal objects. \\
法  & The water-elbow method (basically crying) during a trial is an effective method to get the crowd on your side; just a few of those waterworks and poof, your dissenters are all gone. \\
岩  & In \textit{The Legend of Zelda: Ocarina of Time}, there’s that one giant boulder at the base of Death Mountain that you need to destroy with bombs. \\
炭  & Right below Mt.\ Pyre in \textit{Pokémon Emerald} is the Jagged Pass where you can collect the mountain’s ashes in your soot sack; the wild Pokémon in this area also have a chance of dropping charcoal. \\
掌  & In the outhouse I manipulate my hand to wipe my second mouth clean. (Forgive me for the inappropriate mental image.) \\
波  & One reading of this kanji in Japanese is ``Nami'', which incidentally is a name of a \textit{One Piece} character. Nami always makes sure the ship avoids dangerous waves and also tends to wear many different pelts (if I can stretch the meaning of pelts a little \textbf{:P}). \\
婆  & Link’s grandma (an old woman from \textit{The Wind Waker}) always prays for his safety just beyond the waves. \\
瞬  & ``When I spotted that cute little birdie downtown, out of her birdcage-like university, I did what any cool guy would do: I lifted up my sunglasses, stared her right in the eye, and gave her a wink.'' \\
聖  & A holy man must listen to God’s words. God’s mouth is above that of the king in his perspective. \\
覧  & In \textit{Shin Megami Tensei IV}, when the slave-like “casualties” were given literature, they ended up reclining and reading instead of working. Their perusal angered the luxury class. \\
尽 & Ever heard the sound of a shakuhachi? All you need is an ice-cold drink to go along with it to bring your exhaust down. \\
炉 & In order to keep the fire in the hearth from engulfing us, please keep the door closed! \\
届 & In games, it’s common to find flags sprouted on your map when receiving delivery quests. \\
斥 & There was a knockout tournament at sports today; the team desperately made appeals for the fouls that they accuse the other team of, but these were rejected and the club fell in the rankings until eventually they were axed from the tournaments. \\
遮 & You should watch the road when traveling with a wallet of twenties. Commoners tend to dig in your blind spots for loot and disappear quickly into their caverns. \\
捉 & Pirates are known for nabbing. When you hear the sound of a wooden leg, don’t tremble with your fingers in your mouth—flee. \\
\bottomrule
\end{tabular}
\captionof{table}{Learner cluster 2 representative learner Tactics15 authored mnemonics.}
\label{tab:cluster_2_rep}
\end{minipage}

\begin{minipage}{\textwidth}
\small
\centering
\begin{tabular}{l p{13.5cm}}
\toprule
\textbf{Username} & \textbf{Mnemonic} \\
\midrule
tedperry52 & Ted 亻 was the ninja responsible for putting a hat on the samurai 壬. ("ninja" implies the reading ニン. Other reading: まか・せる.) 責任 (せきにん) duty; responsibility. \\
Puchatek & Mr T scolds an irresponsible porter who lost his luggage, exclaiming, "Where's your responsibility, foo? My favourite golden chain was in that luggage you lost!" \\
elmaestrokgb & Han Solo, a smuggler and porter of goods, takes responsibility for the cargo he carries — hence Jabba's anger. \\
dezts & Jesus had the enormous responsibility of being a porter for humanity's sins, dying for the load he carried. \\
Asayoru & Captain Falcon is tricked by a porter into thinking it’s his own responsibility to carry his luggage. \\
idalton & Captain Picard assigns Data as a porter to carry gear too heavy for biological lifeforms. \\
sherefyounan & A brave person faces a king, declaring: "It is your responsibility to offer us a better life!" \\
eboyj & A porter (壬) is a person (亻) with the responsibility to care for people's luggage. \\
constructionsite & Jesus upheld his responsibility as a porter for his flock. \\
james007123 & It is the Sherpa's responsibility to be the porter and carry everyone's gear on Everest. \\
RandomNumber & A gentleman must take responsibility for the person born from his drop (see 妊 \#546). \\
kariok & Aragorn isn't the right porter for the ring — that responsibility belongs to Frodo. \\
mrnarse & Mr. T, with great power, accepts the responsibility of being a great porter. \\
cloudstrife543 & Mr. T says, ``With great porter, comes great responsibility.'' RTK II: ニン 責任 せきにん (responsibility). \\
Alyangele & A pilot has the responsibility of being the person porter. \\
\bottomrule
\end{tabular}
\captionof{table}{Mnemonics authored by learners for Kanji Cluster 3 representative kanji: 任.}
\label{tab:cluster_3_rep}
\end{minipage}

\newpage
\section{Prompts}
\begin{table}[ht]
  \centering
  \begin{minipage}[t]{0.48\linewidth}
    \centering
    \textbf{EM} \\[6pt]
    \begin{tabular}{@{}p{\linewidth}@{}}
      \toprule
      \textbf{Task Description} \\[6pt]
      Create a single memorable and effective story for the kanji character below. \textbf{You must strictly follow the provided rules when creating the story.} \\[12pt]
      \textbf{Input Data} \\[6pt]
      \textbf{Kanji Character:} \\ \{kanji\} \\[6pt]
      \textbf{Meaning of the Character:}\\  \{meaning\} \\[6pt]
      \textbf{Keywords Representing Its Components:} \\ \{keywords)\} \\[12pt]
      \textbf{Rules for Generating the Story:} \\[6pt]
      \{rules)\}  \\[3pt]
      \textbf{Output (Story):} \\[6pt]
      Please write a story that incorporates the meaning of the kanji and its components, and that \textbf{strictly adheres to the rules provided above}. Do not include any explanations or additional text. \\
      \bottomrule
    \end{tabular}
  \end{minipage}%
  \hfill
  \begin{minipage}[t]{0.48\linewidth}
    \centering
    \textbf{SFT} \\[6pt]
    \begin{tabular}{@{}p{\linewidth}@{}}
      \toprule
      \textbf{Task Description} \\[6pt]
      Create a single memorable and effective story for the kanji character below. You may use your creativity freely to generate the story. \\[12pt]
      \textbf{Input Data} \\[6pt]
      \textbf{Kanji Character:} \\ \{kanji\} \\[6pt]
      \textbf{Meaning of the Character:}\\  \{meaning\} \\[6pt]
      \textbf{Keywords Representing Its Components:} \\ \{keywords)\} \\[12pt]
      \textbf{Output (Story):} \\[6pt]
      Please write a story that incorporates the meaning of the kanji and its components. Do not include any explanations or additional text. \\
      \bottomrule
    \end{tabular}
  \end{minipage}
  \caption{Comparison of the prompt used in our method (left) and the baseline method (right).}
  \label{tab:prompt_for_gen}
\end{table}

\newpage
\section{Evaluation}

\begin{minipage}{\textwidth}
\centering
\begin{tabular}{p{15cm}}
\toprule
\textbf{Task Description} \\
An instruction (might include an Input inside it), two responses to evaluate (denoted as Response A and Response B), a reference answer, and an evaluation criteria are given.

\begin{enumerate}
    \item Write a detailed feedback that assesses the quality of the two responses strictly based on the given evaluation criteria, not evaluating in general.
    \item Make comparisons between Response A, Response B, and the Reference Answer. Instead of examining Response A and Response B separately, go straight to the point and mention the commonalities and differences between them.
    \item After writing the feedback, indicate the better response, either ``A'' or ``B''.
    \item The output format should look as follows: \texttt{Feedback: (write a feedback for criteria) [RESULT] (Either ``A'' or ``B'')}
    \item Please do not generate any other opening, closing, and explanations.
\end{enumerate}

\textbf{Instruction} \\
For the kanji {\{\textit{kanji}\}} meaning {\{\textit{meaning}\}}, which is described by the mnemonic keywords {\{\textit{keywords}\}}, the user wrote: {\{\textit{ground-truth}\}}.

Now, given the kanji {\{\textit{kanji}\}} meaning {\{\textit{meaning}\}}, which is described by the mnemonic keywords {\{\textit{keywords}\}}, what is the most likely story the user would write to memorize this kanji? \\
\\
\textbf{Response A} \\
\{orig\_response\_A\} \\
\\
\textbf{Response B} \\
\{orig\_response\_B\} \\
\\
\textbf{Reference Answer} \\
\{orig\_reference\_answer\} \\
\\
\textbf{Score Rubric} \\
Which response is more likely to have been written by the user, based on the given history of the user? Focus on the user's writing style and sentence structure. \\
\\
\textbf{Feedback} \\
Feedback: (write a feedback for criteria) [RESULT] (Either ``A'' or ``B'') \\
\bottomrule
\label{tab:win_rate_prompt}
\end{tabular}

\end{minipage}

\newpage
\section{Generated Results}

\begin{longtable}{lp{6cm}p{6cm}}
\toprule
\textbf{Kanji} & \textbf{Ground‐Truth} & \textbf{Prediction} \\
\midrule
\endfirsthead

\multicolumn{3}{@{}l}{\textit{Table \thetable\ continued from previous page}} \\
\toprule
\textbf{Kanji} & \textbf{Ground‐Truth} & \textbf{Prediction} \\
\midrule
\endhead

\midrule
\multicolumn{3}{r}{\textit{Continued on next page}} \\
\endfoot

\bottomrule
\endlastfoot

\textcolor{cyan}{忌 (mourn)} &
  The snake mourned over the loss of her one true love. &
  The snake is mourning the loss of its heart. \\
\textcolor{cyan}{将 (leader)} &
  The turtle leader glued vulture claws to his shell as a sign of his position. &
  The vulture leader was so afraid of the turtle that he glued his legs together. \\
\textcolor{cyan}{袋 (sack)} &
  A sack is a poor substitute for a garment. &
  A sack is a garment for carrying things. \\
\textcolor{cyan}{憶 (recollection)} &
  Recollection is the state of recalling ideas from your past. &
  The recollection of the past is a state of mind. \\
\textcolor{cyan}{仲 (go‐between)} &
  Anakin was employed as a go‐between between the Jedi council and Palpatine. &
  The man in black was the go‐between between Han and the Empire. \\
\textcolor{brown}{癖 (mannerism)} &
    X (辟) has many mannerisms/habits, one of them is the ``ok''/``precision'' hand gesture. So when more and more people are dealing with sickness (Covid‐19), X falls on his habit with his ``ok'' gesture... ``It’s gonna be ok, it’s just a flu, it’s gonna be better than ok, it’s gonna be wonderful, it’s gonna disappear like a miracle'' 癖 くせ habit, tendency; 
    口癖 (くちぐせ) catchphrase, manner of talking; 
    癖に (くせに) and yet, though, when, in spite of.
  &
  The mannerism of a person with a sickness is to constantly lick ketchup from their fingers. \\
\textcolor{brown}{描 (sketch)} &
  When learning to SKETCH, it can be good to start with SKETCHing simple still life. Here someone is holding out a FINGER (a thumb, as artists do), to make mental measurements, while SKETCHing the similarly sized SAPLING. (The SKETCH itself grows onto the paper like a FLOWER from your BRAIN, through your FINGERS and onto the paper). &
  A sketch artist is drawing a seedling with her fingers. \\
\textcolor{brown}{渡 (transit)} &
    After getting my (crotch) heated up a few degrees by the (caverns), I wanted to look around, then came to this running water, but the (caverns) clearly continued on the other side… So decided to transit across the water and see what was deeper inside… Man, that water must have been at least 40 °C!! 渡る (わたる) to cross over; 
    渡す (わたす) to hand over; 
    見渡す (みわたす) to look out over; 
    譲渡 (じょうと) transfer, assignment, conveyance; 
    渡航 (とこう) voyage.
  &
  The boatman is the transit from the underworld to the afterlife. \\
\textcolor{brown}{化 (change)} &
  When spooning with your significant other/person, you tend to change positions with one another to get comfortable. &
  The person who changed the world with his spoon was Jesus. \\
\textcolor{brown}{叔 (uncle)} &
  When you’re SMALL and your UNCLEs make more of an impression, they tower ABOVE you, your full height is just at their CROTCH (pictograph of a little figure beside the uncle’s legs). Note: the SMALL signifies this UNCLE is younger than your parent. &
  The uncle is a bit of a weirdo, but he is still a cool uncle. \\
\caption{Sorted by MUD score for predictions, the \textcolor{cyan}{Top-5} are OrangeNut, Culden2, Bagua, Bagua, and SwiftyFrisco, while the \textcolor{brown}{Bottom-5} are Seiun, Seiun, Seiun, nisf01, and Seiun.}
\label{tab:qual_anylsis}
\end{longtable}

\newpage
\section{Ablation Studies}
\subsection{Number of Rules}
\begin{minipage}{\textwidth}
\small
\begin{tabular}{cp{15cm}}
\toprule
\textbf{\#} & \textbf{Rules} \\
\midrule
1 & A direct cause-and-effect story aligning the kanji’s meaning with its components through logical reasoning or functional explanation. \\
2 & Juxtaposing unrelated or contrasting keywords to evoke the kanji's meaning through unexpected connections or contrasts. \\
3 & Forming a straightforward narrative or image that directly associates the keywords with the kanji's meaning, emphasizing simplicity and clarity without additional embellishments. \\
4 & Transforming the keywords into a metaphorical scenario or symbolic representation that evokes the kanji’s meaning through implied transformation or analogy. \\
5 & Using cultural references or idiomatic expressions to leverage shared societal knowledge, enhancing the connection between the keywords and the kanji's meaning through familiarity and relatability. \\
\bottomrule
\end{tabular}
\captionof{table}{K = 5. Learned rules from EM algorithm.}
\label{tab:rules_k_5}
\end{minipage}


\begin{table*}[h]
\small
\begin{tabular}{cp{15cm}}
\toprule
\textbf{\#} & \textbf{Rules} \\
\midrule
1 & Directly identifying the kanji's components as embodying or being synonymous with the kanji's meaning, enhancing memorability through intrinsic identity representation. \\
2 & Illustrating a transformation or change from one state to another, where the keywords embody a process leading to the kanji's meaning, enhancing memorability through the depiction of transition or evolution. \\
3 & Referencing or adapting familiar phrases, idioms, or sayings in a way that connects the kanji's components to its meaning, leveraging linguistic and cultural familiarity to enhance memorability. \\
4 & Structuring the kanji's keywords into a cause-and-effect relationship where the presence or occurrence of the keywords naturally or logically results in the kanji's meaning, enhancing memorability through logical deduction. \\
5 & Crafting a declarative statement that directly asserts the kanji's meaning through the keywords, enhancing memorability through the perception of factual or axiomatic truth. \\
6 & Framing the kanji's components within a universally relatable or anecdotal experience, enhancing memorability through shared human understanding or common life events. \\
7 & Formulating a straightforward factual assertion that connects the kanji's components to its meaning through universally accepted truths or common observations, enhancing memorability through perceived accuracy or realism. \\
8 & Depicting the kanji’s components in a direct, literal enactment or existence that embodies the kanji's meaning, enhancing memorability through tangible visualization. \\
9 & Combining the kanji's components to form a distinct object or artifact that embodies the kanji's meaning, enhancing memorability through tangible creation or representation. \\
10 & Aligning the kanji's keywords with an intuitive or self-evident truth that reflects a clear and straightforward relationship to the kanji's meaning, enhancing memorability through direct and obvious association. \\
11 & Linking the kanji's keywords to its meaning through straightforward, direct statements that emphasize immediate association or relevance, enhancing memorability through clarity and simplicity. \\
12 & Emphasizing a requirement or essential component necessary for achieving the kanji's meaning, enhancing memorability through the depiction of indispensability or necessity. \\
13 & Personifying the kanji's components, attributing roles or actions to them that naturally express the kanji's meaning, enhancing memorability through engaging characterization. \\
14 & Illustrating a natural compatibility or inherent relationship between the kanji's components that seamlessly embodies the kanji's meaning, enhancing memorability through intuitive harmony and cohesion. \\
15 & Constructing a concise narrative that centers around a specific action or outcome involving the kanji's keywords, enhancing memorability through vivid scene depiction and interaction. \\
\bottomrule
\end{tabular}
\caption{K = 15. Learned rules from EM algorithm.}
\label{tab:rules_k_15}
\end{table*}

\end{CJK}
\end{document}